\documentclass[conference, 9pt]{IEEEtran}
\IEEEoverridecommandlockouts
\usepackage{cite}
\usepackage{amsmath,amssymb,amsfonts}
\usepackage{algorithmic}
\usepackage{graphicx}
\usepackage{textcomp}
\usepackage{xcolor}

\usepackage{microtype}
\usepackage{booktabs}
\usepackage{hyperref}
\usepackage{array}
\usepackage{multirow}
\usepackage{algorithm}
\usepackage{algorithmic}
\usepackage{comment}
\usepackage{subcaption}
\usepackage{wrapfig}

\newcommand{\cmtColor}[1]{#1}

\newcommand{\todo}[1]{#1}
\newcommand{\edd}[1]{#1}
\newcommand{\KPP}{KPP}

\def\BibTeX{{\rm B\kern-.05em{\sc i\kern-.025em b}\kern-.08em
    T\kern-.1667em\lower.7ex\hbox{E}\kern-.125emX}}
\begin{document}


\title{{\fontsize{19.45}{0}\selectfont Improving Realistic Worst-Case Performance of NVCiM DNN Accelerators through Training with Right-Censored Gaussian Noise}
}

\author{
    \textbf{
        Zheyu Yan\textsuperscript{$\dagger\circledast$} \ \ \ \ 
        Yifan Qin\textsuperscript{$\dagger$} \ \ \ \ 
        Wujie Wen\textsuperscript{$\ddagger$} \ \ \ \ 
        Xiaobo Sharon Hu\textsuperscript{$\dagger$} \ \ \ \ 
        Yiyu Shi\textsuperscript{$\dagger*$}
    }\\
    \IEEEauthorblockA{\textsuperscript{$\dagger$}University of Notre Dame, \textsuperscript{$\ddagger$}North Carolina State University \{\textsuperscript{$\circledast$}zyan2, \textsuperscript{$*$}yshi4\}@nd.edu}
    \vspace{-1cm}
}

\maketitle

\begin{abstract}

Compute-in-Memory (CiM), built upon non-volatile memory (NVM) devices, is promising for accelerating deep neural networks (DNNs) owing to its in-situ data processing capability and superior energy efficiency. Unfortunately, the well-trained model parameters, after being mapped to NVM devices, can often exhibit large deviations from their intended values due to device variations, resulting in notable performance degradation in these CiM-based DNN accelerators. There exists a long list of solutions to address this issue. However, they mainly focus on improving the mean performance of CiM DNN accelerators. How to guarantee the worst-case performance under the impact of device variations, which is crucial for many safety-critical applications such as self-driving cars, has been far less explored. In this work, we propose to use the k-th percentile performance (\KPP) to capture the realistic worst-case performance of DNN models executing on CiM accelerators.
Through a formal analysis of the properties of \KPP~and 
the noise injection-based DNN training, we demonstrate that injecting a novel right-censored Gaussian noise, as opposed to the conventional Gaussian noise, significantly improves the \KPP~of DNNs.
We further propose an automated method to determine the optimal hyperparameters for injecting this right-censored Gaussian noise during the training process. Our method achieves up to a 26\% improvement in \KPP~compared to the state-of-the-art methods employed to enhance DNN robustness under the impact of device variations.

\end{abstract}

\section{Introductions}

Deep neural networks (DNNs) have demonstrated remarkable advancements, surpassing human performance in a wide range of perception tasks.
The recent emergence of deep learning-based generation models, such as DALL-E~\cite{ramesh2021zero} and the GPT family~\cite{brown2020language}, has further reshaped our workflows. To date, the trend of incorporating on-device intelligence across edge platforms such as mobile phones, watches, and cars, has become an evident~\cite{jiang2020device,chen2016eyeriss,yan2020single}, transforming every walk of life. 
However, the limited computational resources and strict power constraints of these edge platforms present challenges. These circumstances necessitate more energy-efficient DNN hardware beyond the general-purpose CPUs and GPUs. 


Compute-in-Memory (CiM) DNN accelerators~\cite{shafiee2016isaac}, on the other hand, are competitive alternatives to replace CPUs and GPUs in accelerating DNN inference on edge. In contrast to the traditional von Neumann architecture platforms, which involve frequent data movements between memory and computation components, CiM DNN accelerators reduce energy consumption by enabling in-situ computation directly at the storage location of weight data. Moreover, emerging non-volatile memory (NVM) devices, such as ferroelectric field-effect transistors (FeFETs) and resistive random-access memories (RRAMs),
allows NVCiM accelerators to achieve higher memory density and improved energy efficiency compared to conventional MOSFET-based designs~\cite{chen2016eyeriss}. 
However, the reliability of NVM devices can be a concern due to device-to-device (D2D) variations incurred by fabrication defects and cycle-to-cycle (C2C) variations due to thermal, radiation, and other physical impacts. These variations can have a notable negative impact on NVCiM DNN accelerators' inference accuracy, as they may introduce significant differences between the weight values read out from NVM devices during inference and their intended values.

Various strategies have been proposed to mitigate the impact of device variations. These strategies can be broadly categorized into two categories: reducing device value deviations and enhancing the robustness of DNNs in the presence of device variations. 
Device value deviations can be reduced through methods such as write-verify~\cite{shim2020two}, which iteratively applies programming pulses to reduce device value deviation from the desired value after each write.
On the other hand, there exist various approaches that enhance DNN robustness in the presence of device variations. One direction is to identify novel DNN topologies that are more robust in the presence of device variations. This can be achieved through techniques such as neural architecture search~\cite{yan2021uncertainty, yan2022radars} or by leveraging Bayesian Neural Networks~\cite{gao2021bayesian} which use variational training to improve DNN robustness. Another line of methods focuses on training more robust DNN weights using noise injection training~\cite{jiang2020device, he2019noise, yang2022tolerating}. 
In this approach, randomly sampled noise is injected into DNN weights during the forward and backpropagation phases of DNN training.
After the gradient is calculated through backpropagation, the noise is then removed and the weight value without noise is updated by gradient descent. By simulating a noisy inference environment, the noise injection training methods significantly enhance the robustness of DNN models across various DNN topologies.

However, all aforementioned methods 
merely focus on improving the average accuracy performance of CiM DNN accelerators in the presence of device variations, which may be acceptable for non-safety critical applications. In safety-critical applications like airplanes, autonomous driving, and medical devices, even a prediction failure that happens with an extremely low probability (namely worst-case scenario), 
is not affordable because it may result in loss of life, as has been demonstrated in the recent work~\cite{yan2022computing}. 
\todo{The worst-case performance of a DNN model in the presence of device variations can be determined by carefully calibrating the perturbation injected on each weight value to reach the lowest possible DNN performance. Recent work~\cite{yan2022computing} has demonstrated that even a weight value perturbation of less than 3\% can degrade a DNN model's performance to the level of random guessing. However, the likelihood of such a worst-case scenario occurring is extremely low ($<10^{-100}$), which can be safely ignored in common natural environments~\cite{yan2022computing}. Consequently, a more suitable metric to depict the realistic worst-case performance of DNNs in the presence of device variations is needed.}

\todo{To capture realistic worst-case scenarios precisely in the presence of device variations, in this work, we propose to use the k-th percentile performance (\KPP) metric, instead of the average or absolute worst-case performance. With a predetermined $K$ value, the \KPP~metric aims to identify a performance score that the model's performance is consistently greater than this score in all but k\% of cases.\footnote{This research was partially supported by NSF under grants 
CNS-1919167, CCF-2006748, and CCF-2011236, also by ACCESS – AI Chip Center for Emerging Smart Systems, sponsored by InnoHK funding, Hong Kong SAR.}
For example, if a model has a \KPP~of 0.912 when $K=1$, this suggests that the likelihood of a model's performance being greater than 0.912 is 99\% (except the 1\% of the cases). 
When a realistically small $K$ value is given, such as $K=1$, \KPP~can capture a realistic worst-case performance of a DNN model because it (1) guarantees a lower bound of the model's performance and (2) filters out extreme corner cases.
Given the same $K$ value, a higher \KPP~for a DNN model is desirable as it signifies that the model can consistently deliver high performance within a certain probability threshold.}

\todo{Since improving \KPP~guarantees higher realistic worst-case performance of a DNN model,} 
\edd{we revisited the state-of-the-art (SOTA) Gaussian noise injection training method to analyze its effectiveness in improving \KPP. Gaussian noise injection training is widely used simply because it injects noises that statistically mirror the noises in the inference environment. Although it is empirically valid to state that a precise simulation of the inference environment during training would yield optimal results, there is no theoretical proof for it.}
\todo{Thus, \edd{to prove the effectiveness of Gaussian noise injection training}, we thoroughly analyze the relationship between \KPP~of a DNN model and the properties of DNN weights to show what kind of models would provide higher \KPP. Surprisingly, our analysis shows that Gaussian noise injection training is far from optimal in generating robust DNN models in the presence of device variations.}

\todo{Specifically, our key observation is that achieving a higher \KPP~in the presence of device variation needs to satisfy the following three requirements simultaneously: (1) higher DNN accuracy under no device variation; (2) smaller $2^{nd}$ derivatives \emph{w.r.t.} DNN weights, and (3) larger $1^{st}$ derivatives \emph{w.r.t.} DNN weights.
\edd{However}, our analysis (see Section~\ref{sect:train_ana}) shows that the conventional Gaussian noise-injected training approaches can only fulfill the first two requirements, but not the third, making them ineffective for \KPP~improvement. Specifically, the third requirement necessitates distributions with non-zero expected values, a condition that the Gaussian distribution fails to satisfy.}




To this end, we develop TRICE, a method that injects adaptively optimized right-censored Gaussian (RC-Gaussian) noise in the training process. The abbreviation of this method is derived from the name \underline{T}raining with \underline{RI}ght-\underline{C}ensored Gaussian Nois\underline{E} (TRICE), to address all aforementioned three requirements simultaneously.    
TRICE differs from existing approaches in several aspects: (1) rather than using the general Gaussian noise, TRICE uses RC-Gaussian noise which exhibits a unique feature--for all sampled values greater than a designated threshold, the sample value is fixed (\emph{i.e.}, censored) to the threshold. \edd{This results in a negative expected value for the injected noise, thus meeting the third requirement.}
(2) TRICE requires additional hyperparameters tuning, \emph{e.g.}, via a dedicated adaptive training method to identify the optimal noise hyperparameters within a single run of DNN training, which is different from the conventional Gaussian noise-based approaches using the same noise hyperparameters in training and inference. The main contributions of this work are multi-fold: 
\begin{itemize}
    \item We analytically derive the relationship between \KPP~and the gradients of weights and demonstrate how noise injection training can improve \KPP.
    \item We propose to inject right-censored Gaussian noise during DNN training to improve the \KPP~in the presence of device variations. An adaptive training method that can automatically identify optimal noise hyperparameters in the training process is developed accordingly. 
    \item Extensive experimental results show that TRICE improves the $1^{st}$ percentile performance (in terms of top-1 accuracy) in the presence of device variations by up to 15.42\%, 25.09\%, and 26.01\% in LeNet for MNIST, VGG-8 for CIFAR-10 and ResNet-18 for CIFAR-10, respectively compared with SOTA baselines.
    \item We also demonstrate the scalability of our proposed TRICE. That is, in addition to evaluations on uniform RRAM devices, TRICE also improves the $1^{st}$ percentile accuracy by up to 15.61\%, and 12.34\% in two different types of FeFET devices respectively.
    \item To the best of our knowledge, this is the first work that advocates improving \KPP~in NVCiM DNN accelerators with device variations specifically for safety-critical applications. 
\end{itemize}

\section{Related Works}\label{sect:related}

\subsection{Crossbar-based Computing Engine}\label{sec:2.1}

\begin{figure}[ht]
    \vspace{-0.4cm}
    \centering
    \begin{minipage}[b]{0.53\linewidth}
        \includegraphics[trim=10 215 510 10, clip, width=1.\linewidth]{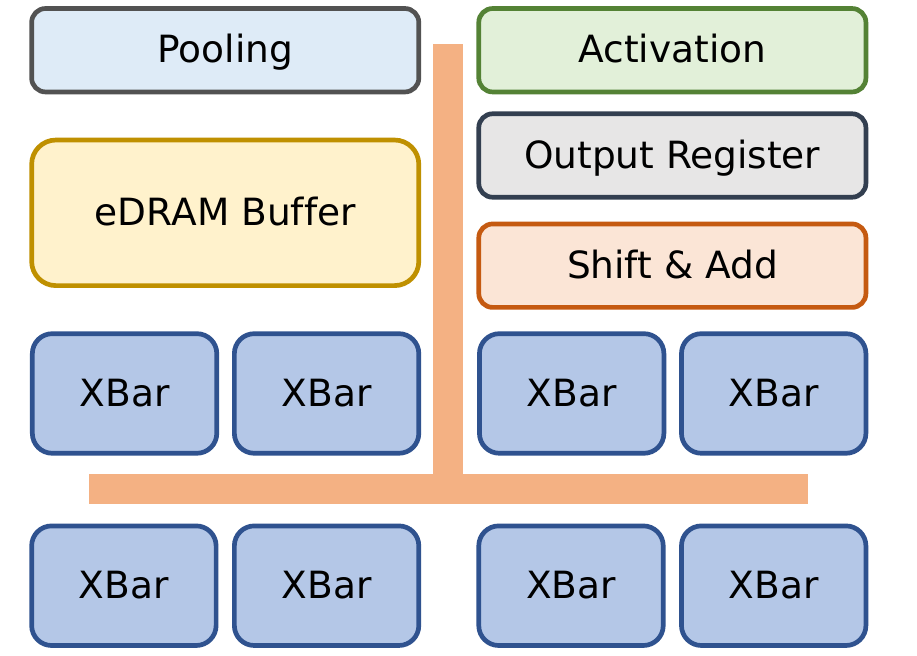}  
        \vspace{-0.5cm}
        \subcaption{NVCiM DNN accelerator overview.}
    \end{minipage}
    \begin{minipage}[b]{0.4\linewidth}
        \includegraphics[trim=0 150 550 0, clip, width=1.\linewidth]{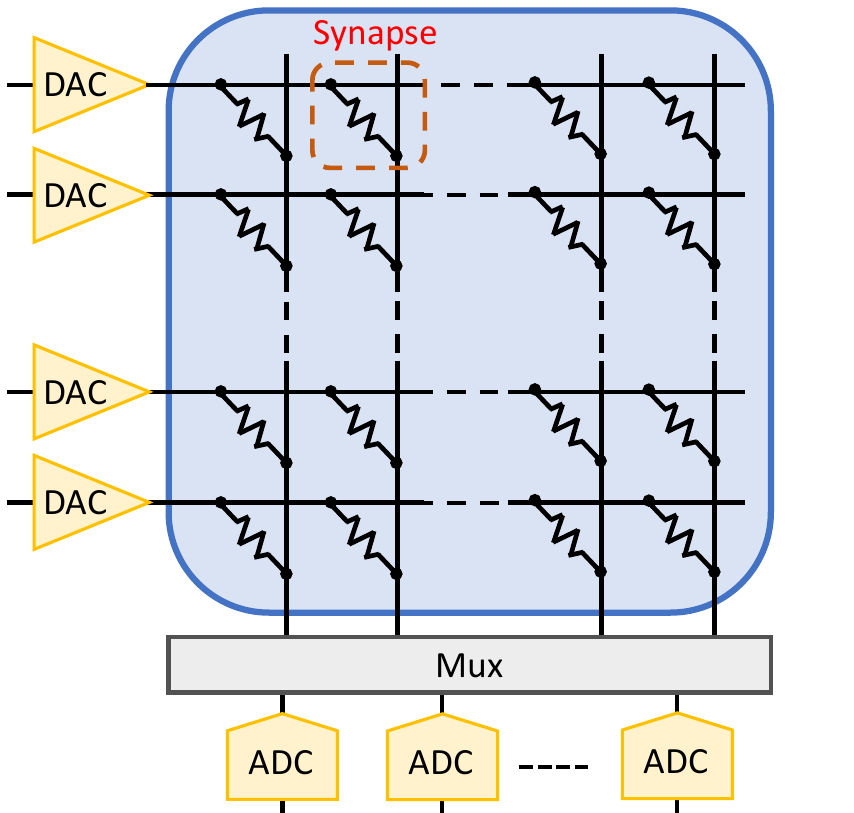}
        \vspace{-0.5cm}
        \subcaption{Crossbar array.}
    \end{minipage}
    \vspace{-0.1cm}
    \caption{Illustration of the NVCiM DNN accelerator architecture for (a) architecture overview and (b) crossbar (XBar) array. In a crossbar array, the input is fed horizontally and multiplied by weights stored in the NVM devices at each cross point. The multiplication results are summed up vertically and the sum serves as an output. The outputs are converted to the digital domain and further processed using digital units such as non-linear activation and pooling.}
    \vspace{-0.3cm}
\end{figure}

The computation engine driving NVCiM DNN accelerators is the crossbar array structure, which can perform matrix-vector multiplication in a single clock cycle. Crossbar arrays store matrix values (\emph{e.g.}, weights in DNNs) at the intersection of vertical and horizontal lines using NVM devices (\emph{e.g.}, RRAMs and FeFETs), while vector values (\emph{e.g.}, inputs for DNNs) are fed through horizontal data lines (word lines) in the form of voltage. The output is then transmitted through vertical lines (bit lines) in the form of current. While the crossbar array performs calculations in the analog domain according to Kirchhoff's laws, peripheral digital circuits are needed for other key DNN operations such as shift \& add, pooling, and non-linear activation. Additional buffers are also needed to store intermediate data. 
Digital-to-analog and analog-to-digital conversions are also needed between components in different domains.


Crossbar arrays based on NVM devices are subject to a number of sources of variations and noise, including spatial and temporal variations. Spatial variations arise from defects that occur during fabrication and can be both local and global in nature. In addition, NVM devices are susceptible to temporal variations that result from stochastic fluctuations in the device material. These variations in conductance can occur when the device is programmed at different times. Unlike spatial variations, temporal variations are usually independent of the device but could be subject to the programmed value~\cite{feinberg2018making}.
For the purpose of this study, we have considered the non-idealities to be uncorrelated among the NVM devices. However, our framework can be adapted to account for other sources of variations with appropriate modifications.

\subsection{Evaluating DNN Robustness in the Presence of Device Variations}

Most existing research uses Monte Carlo (MC) simulations to assess the robustness of NVCiM DNN accelerators in the presence of device variations. This process typically involves extracting a device variation model and a circuit model from physical measurements. The DNN to be evaluated is then mapped onto the circuit model, and the desired value for each NVM device is calculated. In each MC run, one instance of a non-ideal device is randomly sampled from the device variation model, and the actual conductance value of each NVM device is determined. DNN performance (\emph{e.g.}, classification accuracy) in this non-ideal accelerator, is then recorded. This process is repeated numerous times until the collected DNN performance distribution converges. Existing practices~\cite{he2019noise, liu2019fault} generally include around 300 MC runs. This number of MC runs is empirically sufficient according to the central limit theorem~\cite{yan2021uncertainty}.

Only a few researchers are focusing on the worst-case scenarios of NVCiM DNN accelerators in the presence of device variations. A line of research~\cite{wu2020adversarial, tsai2021formalizing, yan2022computing} focuses on determining the worst-case performance by identifying weight perturbation patterns that can cause the most significant decrease in DNN inference performance, while still adhering to the physical bounds of device value deviations. 
One representative work
~\cite{yan2022computing} shows that DNN classification accuracy can drop to random guesses level when adding a less than 3\% perturbation to weights. However, the likelihood of such a worst-case scenario occurring is lower than $<10^{-100}$, which can be safely ignored in common natural environments~\cite{yan2022computing}. Thus, such kinds of worst-case analyses are impractical in terms of accessing the robustness of an NVCiM DNN accelerator.

Thus, in this work, we advocate using k-th percentile performance, a metric that is both practical and precise, for capturing the worst-case performances of a DNN model.

\subsection{Addressing Device Variations}\label{sec:2.3}
Various approaches have been proposed to deal with the issue of device variations in NVCiM DNN accelerators. Here we briefly review the two most common types: enhancing DNN robustness and reducing device variations. 

A common method used to enhance DNN robustness in the presence of device variations is variation-aware training~\cite{jiang2020device,peng2019dnn+,he2019noise,yang2022tolerating}. Also known as noise injection training, the method injects variation to DNN weights in the training process, which can provide a DNN model that is statistically robust in the presence of device variations. In each iteration, in addition to traditional gradient descent, an instance of variation is sampled from a variation distribution and added to the weights in the forward pass. In the backpropagation pass, the same noisy weight and noisy feature maps are used to calculate the gradient of weights in a deterministic and noise-free manner. Once the gradients are collected, this variation is cleared and the variation-free weight is updated according to the previously collected gradients. The details of noise injection training are shown in Alg.~\ref{alg:ni}. Another fashion of training more robust DNN weights is CorrectNet~\cite{eldebiky2022correctnet}. This approach uses a modified Lipschitz constant regularization during DNN training so that the regularized weights are less prone to the impact of device variations. Other approaches include designing more robust DNN architectures~\cite{jiang2020device, yan2021uncertainty, gao2021bayesian} and pruning~\cite{chen2021pruning}.

\begin{algorithm}[b]
\caption{NoiseTrain~($\mathcal{M}$, $\mathbf{w}$, $\mathcal{D}ist$, $ep$, $\mathbf{D}$, $\alpha$)}
\begin{algorithmic}[1]\label{alg:ni}
\STATE \cmtColor{// INPUT: DNN topology $\mathcal{M}$, DNN weight $\mathbf{w}$, noise distribution $\mathcal{D}ist$, \# of training epochs $ep$, dataset $\mathbf{D}$, learning rate $\alpha$;}

\FOR{($i=0$; $i < ep$; $i++$)}
    \FOR{$x$, $GT$ in $\mathbf{D}$}
        \STATE Sample $\Delta\mathbf{w}_i$ from $\mathcal{D}ist$;
        \STATE $loss = $ CrossEntropyLoss($\mathcal{M}(\mathbf{w}+\Delta\mathbf{w}_i, x)$, $GT$);
        \STATE $\mathbf{w} = \mathbf{w} - \alpha \frac{\partial loss}{\partial \mathbf{w}+\Delta\mathbf{w}_i}$
    \ENDFOR
\ENDFOR
\end{algorithmic}
\end{algorithm}

To reduce device variations induced device value deviation, write-verify\cite{shim2020two, yao2020fully} is commonly used during the programming process. 
\todo{An NVM device is first programmed to an initial state using a pre-defined pulse pattern. Then the value of the device is read out to verify if its conductance falls within a certain margin from the desired value (\emph{i.e.}, if its value is precise). If not, an additional update pulse is applied, aiming to bring the device conductance closer to the desired one. This process is repeated until the difference between the value programmed into the device and the desired value is acceptable. }
This approach is highly effective in reducing the device value deviations, but the process typically requires a few iterations, which is time-consuming. 
There are also various circuit design efforts~\cite{shin2021fault, jeong2022variation} that try to 
mitigate the device variations. 


\section{Proposed Method}\label{sect:proposed}


\todo{In this section, we introduce a novel variant of the noise injection training method designed to improve the k-th percentile performance (\KPP) of a DNN model.
\edd{The conventional noise injection training injects Gaussian noise in the training process simply because \todo{it mirrors the impact of device variations occurring in inference.} There is no theoretical proof that such practice would offer the most robust DNN models. In this section, we show through mathematical analysis that Gaussian noise injection training is far from optimal in improving \KPP.}} Specifically, this section begins with a formal definition of \KPP~and an analysis of its relationship with DNN weights. Next, we analyze the noise injection training framework and identify the requirements for the noise injected during training. We show that Gaussian noise does not satisfy all requirements.

Thus, we propose several candidate noise types and select right-censored Gaussian noise through experimentation. Moreover, we develop an adaptive training method that automatically determines the optimal hyperparameters for the right-censored Gaussian noise injection. The resulting framework is called \underline{T}raining with \underline{RI}ght-\underline{C}ensored Gaussian Nois\underline{E} (TRICE).

\subsection{K-th Percentile Performance}

The \KPP~of a DNN model is derived from the k-th percentile of a distribution. The k-th percentile of a distribution can be defined as the value $z_{pk}$ that separates the lowest k\% of the observations from the highest (100-k)\% of the observations in a distribution. Formally speaking, given a random variable $Z$ following a distribution $\mathcal{D}ist$, there exists a value $z_{pk}$ that, if sampling a value $z_i$ from $Z$, there is a k\% probability that $z_i \leq z_{pk}$. It is equivalent to:
\vspace{-0.1cm}
\begin{equation}
    \vspace{-0.1cm}
    \begin{aligned}
        k/100 = cdf_{\mathcal{D}ist}(z_{pk})
    \end{aligned}
\end{equation}
where $cdf_{\mathcal{D}ist}$ is the cumulative distribution function of $\mathcal{D}ist$. 

\begin{figure}[ht]
\vspace{-0.4cm}
\begin{center}
\centerline{\includegraphics[trim=25 85 40 110, clip, width=1\linewidth] {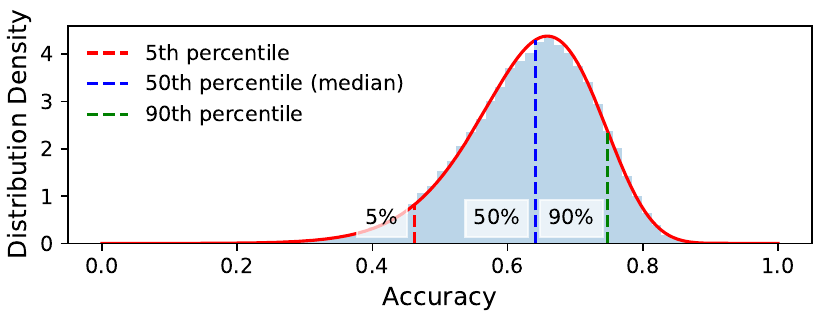}}
\vspace{-0.2cm}
\caption{Illustration of \KPP~(in terms of top-1 accuracy). The red curve represents the accuracy distribution of a DNN in the presence of device variations. The intersection point of each straight line and the x-axis represents the k-th percentile accuracy.}
\label{fig:kthp}
\end{center}
\vspace{-0.6cm}
\end{figure}

In the context of a DNN model's performance in the presence of device variations, the \KPP~represents the minimum performance level that the model achieves with a probability of at least (100-k)\%. For example, As shown in Fig.~\ref{fig:kthp}, the $5^{th}$ percentile performance in terms of top-1 accuracy (\emph{i.e.}, k-th percentile accuracy) of this DNN model in the presence of device variations is 0.4623 which means for 5\% of the cases the DNN accuracy will be lower than 0.4623, and for 95\% of the cases, the DNN accuracy is greater than 0.4623.

\KPP~of a DNN model can be easily evaluated through Monte-Carlo simulation. Specifically, with $N_{sample}$ Monte Carlo runs, $N_{sample}$ performance values are collected. These performance values are then sorted in ascending order and the $(N_{sample} \times k\%)^{th}$ element of this sorted array is the estimation of \KPP. The overall process is shown in Algorithm~\ref{alg:qe}.

\begin{algorithm}[t]
\caption{QuantEval~($\mathcal{M}$, $\mathbf{w}$, $\sigma_d$ $q$, $\mathbf{D}$, $N_{sample}$)}
\begin{algorithmic}[1]\label{alg:qe}

\STATE \cmtColor{// INPUT: DNN topology $\mathcal{M}$, DNN weight $\mathbf{w}$, device value variation $\sigma_d$, $q = k/100$ for k-th percentile, evaluation dataset $\mathbf{D}$, number of samples $N_{sample}$;}
\cmtColor{\STATE // OUTPUT: k-th percentile performance of  $\mathcal{M}(\mathbf{w})$;}
\STATE initialize empty list perf$_l$;
\FOR{($i=0$; $i < N_{sample}$; $i++$)}
    \STATE Sample $\Delta\mathbf{w}_i$ from Gaussian($0$, $\sigma_d$);
    \STATE perf$_i = $ performance of $\mathcal{M}(\mathbf{w} + \Delta\mathbf{w}_i)$ in dataset $\mathbf{D}$;
    \STATE Add value perf$_i$ to list perf$_l$;
\ENDFOR
\STATE perf$_l$ = sort(perf$_l$);
\STATE perf$_q$ = perf$_l[q\times len($perf$_l)]$
\STATE return perf$_q$;
\end{algorithmic}
\end{algorithm}

\vspace{-0.2cm}
\subsection{Relationship Between Weights and k-th Percentile Performance}\label{sect:percentile}
\todo{After establishing the definition of the \KPP, we proceed to analyze how it relates to the trained weights of the DNN model. We use the loss function as the metric for assessing the performance throughout this analysis.}

Given a neural network model $\mathcal{M}$ and its trained weight vector $\mathbf{w}$, the output $\mathbf{out}$ of this model from the input $\mathbf{x}$ can be described as $\mathbf{out} = \mathcal{M}(\mathbf{w}, \mathbf{x})$. Further given the ground truth label $\mathbf{GT}$ and the loss function $f$, its loss can be described as $loss = f(\mathcal{M}(\mathbf{w}, \mathbf{x}), \mathbf{GT})$. Because the values of $\mathbf{x}$ and $\mathbf{GT}$ are fixed when inferencing on a given dataset, the loss expression can be simplified as a function of $\mathbf{w}$, \emph{i.e.}, $loss = f(\mathbf{w})$.

Here we study the impact of perturbing one element $w_0$ in the weight vector $\mathbf{w}$. Specifically, because this weight value is subjected to the impact of device variations, it is perturbed to $w_0 + \Delta w$, where $\Delta{w}$ is the device variation-induced perturbation. We can then apply Taylor expansions to the loss function \emph{w.r.t.} the perturbed weight:
\vspace{-0.1cm}
{\small
\begin{equation}
    \vspace{-0.1cm}
    \begin{aligned}
        f(w_0 + \Delta w) = & f(w_0) + f'(w_0)\Delta{w} + \frac{f''(w_0)}{2} (\Delta{w})^2 + o((\Delta{w})^3) \\
        \approx & f(w_0) + f'(w_0) \Delta{w} + \frac{f''(w_0)}{2} (\Delta{w})^2
    \end{aligned}\label{eq:loss_taylor}
\end{equation}
}

We can observe in Eq.~\ref{eq:loss_taylor} that the loss function can be approximated by a quadratic function of $\Delta{w}$. Given that the weight perturbation $\Delta{w}$ follows the distribution of device variations ($\Delta{w}\sim\mathcal{D}ist$), we can calculate the k-th percentile of the loss as follows:

First, let $q = k/100$ be the probability number of k-th percentile. We then let the unknown k-th percentile be $loss_q$. According to the property of quadratic functions, along with the fact that $f''(w) \geq 0$~\cite{yan2022swim} and $loss_q$ is greater than the minimum value of Eq.~\ref{eq:loss_taylor}, we know that there exist two real numbers $\Delta{w_1}$ and $\Delta{w_2}$, $\Delta{w_1} < \Delta{w_2}$, such that if $\Delta{w_1} < \Delta{w} < \Delta{w_2}$, then $f(\Delta{w}) < loss_q$.

By the definition of \KPP~and the loss is the lower the better, we have $q$ as the probability of $f(\Delta{w}) \geq loss_q$, and then $1-q$ is the probability of $\Delta{w_1} \leq \Delta{w} \leq \Delta{w_2}$. Recalling that weight perturbation $\Delta{w}$ follows the device variation distribution ($\Delta{w}\sim\mathcal{D}ist$), we have:
\vspace{-0.1cm}
\begin{equation}
    \vspace{-0.1cm}
    \label{eq:quantile}
    1 - q = cdf_{\mathcal{D}ist}(w_2) - cdf_{\mathcal{D}ist}(w_1)
\end{equation}
where $cdf_{\mathcal{D}ist}$ is the cumulative distribution function (CDF) of $\mathcal{D}ist$. Through the definition of $w_1$, $w_2$ and $loss_q$, we also know that:
\vspace{-0.1cm}
\begin{equation}
    \vspace{-0.1cm}
    \begin{aligned}
        w_1 &= \frac{-f'(w_0) - \beta}{f''(w_0)}\\
        w_2 &= \frac{-f'(w_0) + \beta}{f''(w_0)}\\
        \beta &= \sqrt{f'(w_0)^2 - 2f''(w_0)(f(w_0)-loss_q)}
    \end{aligned}\label{eq:quadra}
\end{equation}

Combining Eq.~\ref{eq:quantile} and Eq.~\ref{eq:quadra}, we can get an analytical relationship between $q$ and $loss_q$ and thus can calculate $loss_q$ given the device value deviation distribution $\mathcal{D}ist$ and the trained model weight $w_0$.


In this work, we target a device model that the device value deviation follows Gaussian distribution $\mathcal{N}(0,\sigma_d)$, whose CDF is:
\vspace{-0.1cm}
\begin{equation}
    \vspace{-0.1cm}
    \begin{aligned}
        cdf_{\mathcal{D}ist}(w) = \int_{-\infty}^w e^{-t^2} dt
    \end{aligned}\label{eq:cdf}
\end{equation}

Combining Eq.~\ref{eq:quantile} and Eq.~\ref{eq:quadra} and the first-order approximation of Eq.~\ref{eq:cdf}, we obtain:
\vspace{-0.1cm}
\begin{equation}
    \vspace{-0.1cm}
    \begin{aligned}
        loss_q = -\frac{f'(w_0)^2}{2f''(w_0)} + f(w_0) + \frac{f''(w_0)\pi q^2 \sigma_d^2}{4}
    \end{aligned}\label{eq:lossq}
\end{equation}

Considering $f'(w_0)$ as a variable, it is clear that $loss_q$ is a quadratic function \emph{w.r.t.} $f'(w_0)$. Extensive research works~\cite{dangel2020backpack, yan2022swim} have shown that when using cross-entropy loss with softmax as the loss function, the second derivatives of weights \emph{w.r.t.} the loss is positive, \emph{i.e.}, $f''(w_0) > 0$. Thus, it is clear that Eq.~\ref{eq:lossq} reaches its maximum value when $f'(w_0)=0$ and decreases when $f'(w_0)$ diverges from $0$. Therefore, by observing the first term of~\ref{eq:lossq}, to gain a low enough $loss_q$, hence high enough \KPP, a smaller $f''(w_0)$, and a $f'(w_0)$ with larger absolute values is required. Similarly, by observing the second and the third term of~\ref{eq:lossq}, a smaller $f(w_0)$, and a smaller $f''(w_0)$ is required. Thus, to improve the \KPP~of a DNN model, the DNN training process needs to simultaneously minimize $f(w_0)$ and $f''(w_0)$, and maximize $|f'(w_0)|$.

\subsection{The Effect of Noise Injection Training}\label{sect:train_ana}
\todo{According to the conclusion in Section~\ref{sect:percentile}, the DNN training process needs to minimize $f(w_0)$ and $f''(w_0)$, then maximize $|f'(w_0)|$ at the same time. We now analyze the noise injection training process to see how to satisfy these requirements.}


Using similar denotations as Section~\ref{sect:percentile} and recall Alg.~\ref{alg:ni}, one iteration of the noise injection training process can be depicted as:
\vspace{-0.1cm}
\begin{equation}
    \vspace{-0.1cm}
    \begin{aligned}
        w_{t+1} & = w_t - \alpha f'(w_t + \Delta{w})
    \end{aligned}\label{eq:gdoverview}
\end{equation}
where $w_{t}$ is the current weight value, $w_{t+1}$ is the updated weight value after this iteration of training and $\alpha$ is the learning rate.
By applying Taylor expansion on $f'(w_t + \Delta{w})$, we obtain:

\vspace{-0.1cm}
\begin{equation}
    \vspace{-0.1cm}
    \begin{aligned}
        w_{t+1} & \approx w_t - \alpha \left(f'(w_t) + \Delta{w} f''(w_t) + \frac{(\Delta{w})^2}{2}f'''(w_t)\right)
    \end{aligned}\label{eq:wu_overview}
\end{equation}

Considering a noise injection training process where in each iteration of training, the device variation-induced weight value perturbation $\Delta{w}$ is sampled for enough instances instead of only once, the statistical behavior for such noise injection training is:
\vspace{-0.1cm}
\begin{equation}
    \vspace{-0.1cm}
    \begin{aligned}
        & w_{t+1} = w_t - \alpha E_{ \Delta{w}}[f'(w_t + \Delta{w})]\\
                & \approx w_t - \alpha \left(f'(w_t) + E[\Delta{w}] f''(w_t) + \frac{E[(\Delta{w})^2]}{2}f'''(w_t)\right)
    \end{aligned}\label{eq:weight_update}
\end{equation}
where $E[\Delta{w}]$ is the expected value (\emph{i.e.}, mean) of $\Delta{w}$.

By observing Eq.~\ref{eq:weight_update} and by recalling the requirements derived through Eq.~\ref{eq:lossq} that the DNN training process needs to (1) minimize $f(w_0)$, (2) minimize $f''(w_0)$, then (3) maximize $|f'(w_0)|$ at the same time. We can analyze the three terms after $\alpha$ in Eq.~\ref{eq:weight_update} to design the noise distribution to be injected.

For the three terms after $\alpha$, the first term $f'(w_t)$ is the first-order gradient that is used in vanilla gradient descent that minimizes the value of $f(w_{t+1})$. This satisfies the first requirement gained in Section~\ref{sect:percentile}. Another side effect is that, when the training process is close to converging, this term would push the first-order gradient toward zero.

The third term $\frac{E[(\Delta{w})^2]}{2}f'''(w_t)$ affects the second derivatives. Because $E[(\Delta{w})^2]$ is always positive, this term minimizes the value of $f''(w_{t+1})$. This satisfies the second requirement gained in Section~\ref{sect:percentile}.

For the second term $E[\Delta{w}] f''(w_t)$, it affects the first derivatives. As $E[\Delta{w}]$ can be either positive, zero, or negative, this term would respectively minimize, not change, or maximize the first-order gradient. Combined with the first term that pushes the first-order gradient towards zero, injecting a noise with a negative mean would result in a maximized positive first-order gradient and vice versa. Because Eq.~\ref{eq:lossq} requires a first-order gradient of larger absolute value, a noise distribution with a non-zero mean value is required. The widely used Gaussian distribution, whose mean value is zero, however, does not meet this requirement. Therefore, a new type of noise needs to be utilized for noise injection training.

\subsection{Candidate Noise Distributions}
\begin{figure}[ht]
    \centering
    \includegraphics[trim=0 0 0 0, clip, width=0.6\linewidth]{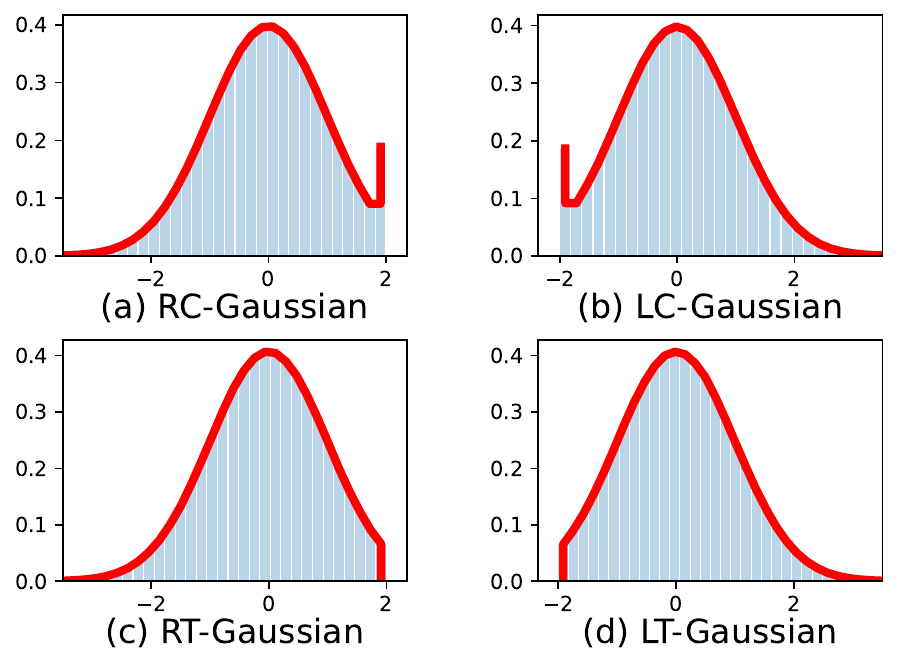}
    \vspace{-0.2cm}
    \caption{The distribution histogram of different candidate noise with $\sigma_t = 1$ and $th=2$. The x-axis represents the perturbation magnitude and the y-axis represents the distribution density.}
    \label{fig:PT}
    \vspace{-0.5cm}
\end{figure}

According to Section~\ref{sect:train_ana}, to improve the model robustness, the distribution injected in the training process needs to satisfy requirements that: $E[(\Delta{w})^2] > 0$ and $E[(\Delta{w})] \neq 0$. We also need this distribution to yield a model with high enough accuracy when noise-free, according to Section~\ref{sect:percentile}.
We propose to consider four candidate noise distributions for our study, all of which are variations of the Gaussian distribution. These distributions include (a) Right-Censored Gaussian (RC-Gaussian), (b) Left-Censored Gaussian (LC-Gaussian), (c) Right-Truncated Gaussian (RT-Gaussian), and (d) Left-Truncated Gaussian (LT-Gaussian). In a Right-Censored Gaussian distribution, all values follow Gaussian distribution except that those greater than a certain threshold are set (censored) to be the threshold value. This applies similarly to the LC-Gaussian distribution except that the value smaller than the threshold is censored. The property of RC-Gaussian is shown in Eq.~\ref{eq:cr-gaussian}. Different from RC and LC-Gaussian, in the Right-Truncated Gaussian distribution, any value greater than a threshold is cut off, which means there is zero probability for the perturbation value to be greater than the threshold. This applies similarly to LT-Gaussian. The distribution histograms of the four candidates are shown in Fig.~\ref{fig:PT}.

\vspace{-0.4cm}
\begin{equation}
    \vspace{-0.1cm}
    \begin{aligned}
        \text{\emph{RC-Gaussian}}(th, \sigma_t) &= \begin{cases}
                th \times \sigma_t,   &\emph{if} \  g \geq th \times \sigma_t \\
                g,   &\emph{else}
            \end{cases}& \\
        g &\sim \mathcal{N}(0, \sigma_t) 
    \end{aligned}\label{eq:cr-gaussian}
\end{equation}

With excessive experiments, we select to inject Right-Censored Gaussian distribution during noise injection training because it would result in the best \KPP. The results of this study are shown in the experiment section.


\subsection{Automated Hyperparameter Selection through Adaptive Training}

\begin{figure}[ht]
    \vspace{-0.4cm}
    \centering
    \includegraphics[trim=90 20 30 50, clip, width=0.7\linewidth]{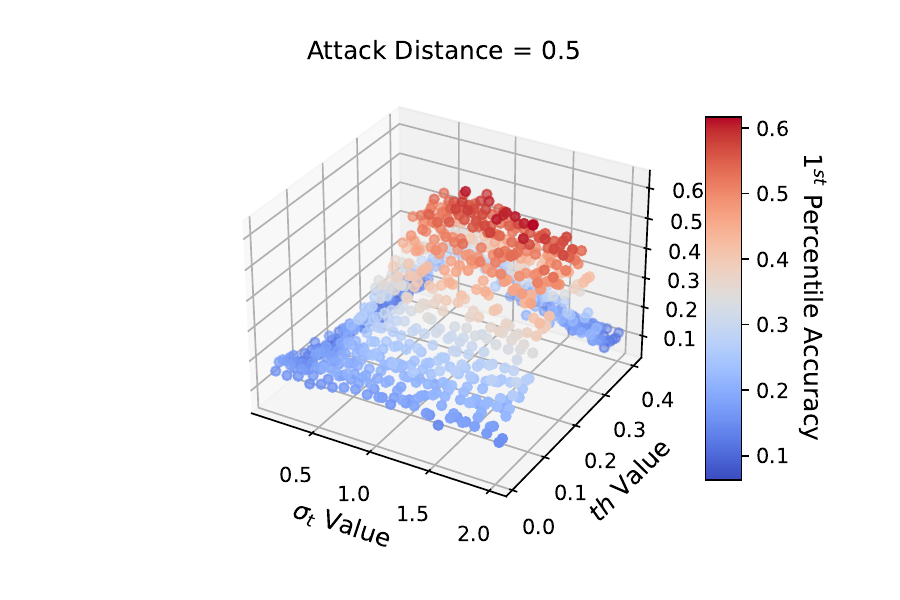}
    \vspace{-0.3cm}
    \caption{Results for the grid search of injecting right-censored Gaussian noise with different hyperparameters on model LeNet for dataset MNIST. The x-axis and y-axis represent the different choices of hyperparameter $\sigma_t$ and $th$, respectively. The z-axis represents the $1^{st}$ percentile accuracy of the trained model. It is clear that the optimal solution sets in the middle of the search space for each hyperparameter.}
    \label{fig:Grid}
    \vspace{-0.3cm}
\end{figure}

Right-Censored Gaussian noise injection training requires massive hyperparameter tuning. \todo{Unlike traditional Gaussian noise injection training, which employs noise hyperparameters the same as the device variation-induced weight value deviation during training to accurately replicate the inference environment, injecting RC-Gaussian noise introduces different types of noise during training and inference.} Thus, the two hyperparameters, $\sigma_t$ and $th$, need to be calibrated for each different DNN model and $\sigma_d$ value.
The process of determining the optimal hyperparameters can be time-consuming and requires significant human effort. AutoML~\cite{yan2022radars}-based methods are possible solutions but they typically require multiple trials to determine the optimal hyperparameter. Therefore, we propose an adaptive training method to find the optimal noise hyperparameters during the training process. This method requires no hyperparameter tuning and takes only one single training run to train the optimal model. To develop this method, we first conduct a grid search of hyperparameters. As shown in Fig.~\ref{fig:Grid}, for both hyperparameters ($\sigma_t$ and $th$), as the value of the hyperparameter increases, the DNN performance initially increases and then decreases after reaching an optimal point. This property allows us to use a binary search-like method to find the optimal hyperparameter values.

Specifically speaking, during the training process, three identical DNN models are initialized and trained simultaneously. Each DNN model is trained by injecting noises with different hyperparameters. The hyperparameters each model uses are determined by the binary search engine. After each epoch, the \KPP~of each model trained under noises with different hyperparameters is evaluated. The binary search engine then updates the hyperparameters of each model according to their performance rankings. The weight of each model is also reassigned by the model that has the highest \KPP. To stabilize training, the model is first trained by $warm$ warm-up epochs without updating the noise hyperparameters. Moreover, to accelerate training, when the binary search method converges, which means all three models are using the same noise hyperparameters, the three models are merged into one model, which means only one model needs to be trained.

The binary search-like policy to identify the optimal value of one hyperparameter is as follows: 
with the starting point $start$ and the ending point $end$, in each iteration, the three candidate values are the three quartiles of $start$ and $end$, \emph{i.e.},  $left = start + 1 \times (end - start)/4$, $mid = start + 2\times(end - start)/4$ and $right = start + 3\times(end - start)/4$. If the model trained with hyperparameter $mid$ has the highest \KPP, this means the optimal value is not in the range of $[start, left]$ and $[right, end]$, so we can perform $start \leftarrow left$ and $end \leftarrow right$. Similarly, if the model trained with hyperparameter $left$ has the highest \KPP, we only perform $end \leftarrow right$, and if the model trained with hyperparameter $right$ has the highest \KPP, we only perform $start \leftarrow left$. This process is performed iteratively until $|end - start|\leq 1e-4$.

\setlength{\textfloatsep}{0.5\baselineskip plus  0.2\baselineskip minus 0.4\baselineskip}
\begin{algorithm}[h]
\caption{TRICE~($\mathcal{M}$, $start$, $end$, $th$, $ep$, $warm$, $N_{train}$, $\sigma_d$, $q$, $\mathbf{D}$, $\alpha$)}
\begin{algorithmic}[1]\label{alg:adpt}
\STATE \cmtColor{// INPUT: DNN topology $\mathcal{M}$, start and end perturbation magnitude $start$, $end$, RC-Gaussian threshold $th$, number of training epochs $ep$, number of warm up epochs $warm$, number of evaluation samples during training $N_{train}$, target device value variation $\sigma_d$, target percentile $q$, dataset $\mathbf{D}$ and learning rate $\alpha$;}

\STATE Initialize three DNN models $\mathcal{M}(\mathbf{w_1})$, $\mathcal{M}(\mathbf{w_2})$, $\mathcal{M}(\mathbf{w_3})$ of topology $\mathcal{M}$;
\FOR{($i=0$; $i < ep$; $i++$)}
    \IF{$end - start < 1e-4$}
        \STATE \cmtColor{// Train only one model when $start == end$.}
        \STATE NoiseTrain($\mathcal{M}$, $\mathbf{w_1}$, RC-Gauss($th$, $start$), 1, $\mathbf{D}$, $\alpha$);
    \ELSE
        \STATE \cmtColor{// Train three models with three different hyperparameters.}
        \STATE $left$ = $start + 1 \times (end - start)/4$;
        \STATE $mid$ = $start + 2 \times (end - start)/4$;
        \STATE $right$ = $start + 3 \times (end - start)/4$;
        \STATE NoiseTrain($\mathcal{M}$, $\mathbf{w_1}$, RC-Gauss($th$, $left$), 1, $\mathbf{D}$, $\alpha$);
        \STATE NoiseTrain($\mathcal{M}$, $\mathbf{w_2}$, RC-Gauss($th$, $mid$), 1, $\mathbf{D}$, $\alpha$);
        \STATE NoiseTrain($\mathcal{M}$, $\mathbf{w_3}$, RC-Gauss($th$, $right$), 1, $\mathbf{D}$, $\alpha$);
        \IF{$i \geq warm$}
            \STATE \cmtColor{// Only evaluate performance and update hyperparameters after warmup.}
            \STATE perf$_1$ = QuantileEval($\mathcal{M}$, $\mathbf{w_1}$, $\sigma_d$, $q$, $\mathbf{D}$, $N_{train}$);
            \STATE perf$_2$ = QuantileEval($\mathcal{M}$, $\mathbf{w_2}$, $\sigma_d$, $q$, $\mathbf{D}$, $N_{train}$);
            \STATE perf$_3$ = QuantileEval($\mathcal{M}$, $\mathbf{w_3}$, $\sigma_d$, $q$, $\mathbf{D}$, $N_{train}$);
            \STATE \cmtColor{// use binary search to update hyperparameters}
            \IF{$\max($perf$_1$, perf$_2$, perf$_3) ==$ perf$_2$}
                \STATE $start, end, \mathbf{w_1}, \mathbf{w_3} = left, right, \mathbf{w_2}, \mathbf{w_2}$;
            \ELSIF{$\max($perf$_1$, perf$_2$, perf$_3) ==$ perf$_1$}
                \STATE $end, \mathbf{w_2}, \mathbf{w_3} = right, \mathbf{w_1}, \mathbf{w_1}$
            \ELSIF{$\max($perf$_1$, perf$_2$, perf$_3) ==$ perf$_3$}
                \STATE $start, \mathbf{w_1}, \mathbf{w_2} = left, \mathbf{w_3}, \mathbf{w_3}$
            \ENDIF
        \ENDIF
    \ENDIF
    
\ENDFOR
\end{algorithmic}
\end{algorithm}

\todo{Note that there are more efficient hyperparameter tuning algorithms available compared to our method. The optimal solution requires training fewer models using different hyperparameters. However, our approach is better suited for noise injection training due to the following reasons. (1) It involves more estimations of model performances using different hyperparameters, thereby reducing the impact of imperfect \KPP~estimations obtained from a small number of Monte Carlo runs. (2) It continuously trains a model using a hyperparameter of $mid = (start + end)/2$, which is closer to the final optimal hyperparameter. This makes the training process easier to converge.}

In our practice, we use adaptive search to automatically find perturbation scale $\sigma_t$ and manually determine $th$.

The whole training framework with automated hyperparameter tuning is named \underline{T}raining with \underline{RI}ght-\underline{C}ensored Gaussian Nois\underline{E} (TRICE) and shown in Algorithm~\ref{alg:adpt}.

\setlength{\textfloatsep}{1.55\baselineskip plus  0.2\baselineskip minus 0.4\baselineskip}




\section{Experimental Evaluation}\label{sect:exp}

In this section, we comprehensively evaluate our proposed TRICE method in terms of \KPP~improvement for CiM DNN accelerators suffering from device variations.
We first discuss how to link the device value variations to additive noise on weights based on the noise model. We then compare the effectiveness of TRICE against SOTA baselines using different datasets, models, and different types of NVM devices that can be used to build NVCiM DNN accelerators. Ablation studies that show the advantages of RC-Gaussian noise over different noise candidates are also conducted.

\begin{figure*}
    \vspace{0.1cm}
    \centering
    \begin{minipage}[b]{0.44\linewidth}
        \includegraphics[trim=10 35 40 60, clip, width=1.\linewidth]{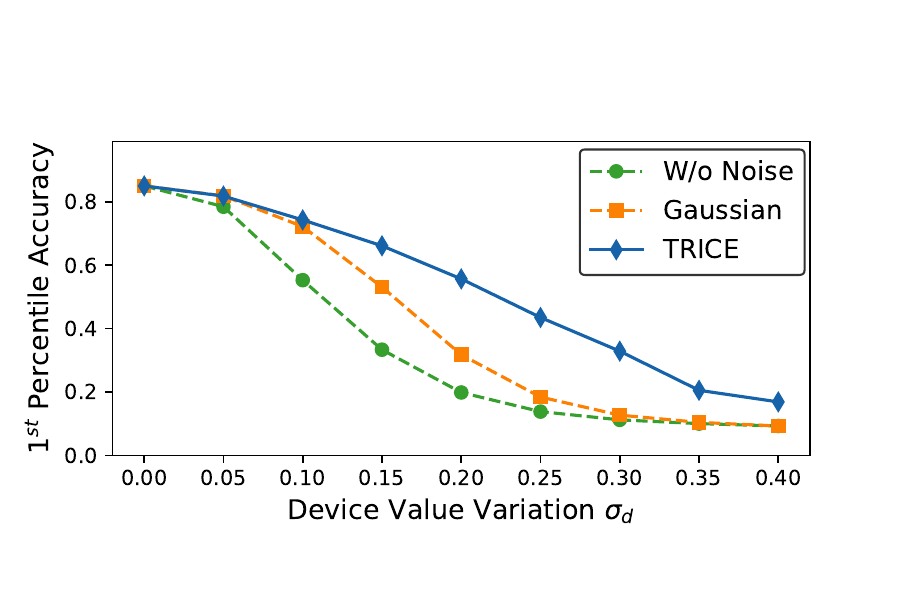}
        \vspace{-0.5cm}
        \subcaption{VGG-8 for CIFAR-10}
        \label{fig:cifar}
    \end{minipage}
    \hspace{0.03\linewidth}
    \begin{minipage}[b]{0.44\linewidth}
        \includegraphics[trim=10 35 40 60, clip, width=1.\linewidth]{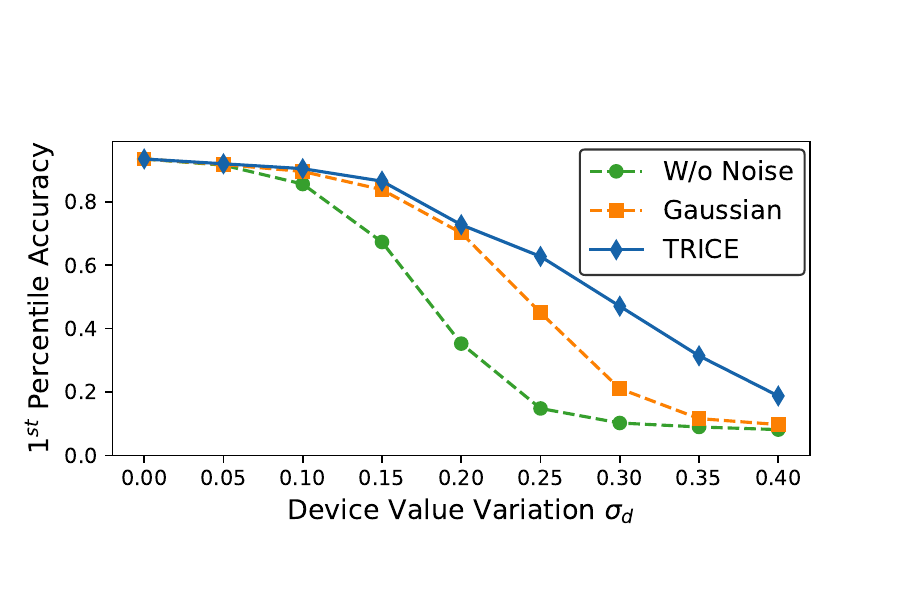}  
        \vspace{-0.5cm}
        \subcaption{ResNet-18 for CIFAR-10}
        \label{fig:res18}
    \end{minipage}
    \vspace{-0.1cm}
    \caption{Comparison of the $1^{st}$ percentile accuracy achieved by models trained using TRICE and baseline methods on (a) VGG-8 and (b) ResNet-18 for dataset CIFAR-10. The x-axis represents the magnitude of device value variation ($\sigma_d$) and the y-axis represents the $1^{st}$ percentile accuracy.}
    \vspace{-0.7cm}
\end{figure*}

\subsection{Modeling of Device Variation-induced Weight Perturbation}
\label{sect:model}
Without loss of generality, we mainly focus on device variations originating from the programming process, in which the conductance value programmed to NVM devices can deviate from the desired value. Next, we show how to model the impact of device variations on DNN weights. 

Assume a $H$ bits DNN weight, the desired weight value after quantization ($\mathcal{W}_{des}$) can be represented as: 

\vspace{-0.2cm}
\begin{equation}
    \vspace{-0.1cm}
    \mathcal{W}_{des} = \frac{\max{|\mathcal{W}|}}{2^H - 1}\sum_{j=0}^{H-1}{h_j \times 2^j}
\end{equation}
where $h_j \in \{0,1\}$ is the value of the $j^{th}$ bit of the desired weight value, $\mathcal{W}$ is the floating point weight value and $\max{|\mathcal{W}|}$ is the maximum absolute value of the weight. 
For an NVM device capable of representing $B$ bits of data, since each weight value can be represented by $H/B$ devices\footnote{Without loss of generality, we assume that $H$ is a multiple of $B$.}, the corresponding mapping process can be expressed as: 
\vspace{-0.3cm}
\begin{equation}
    \vspace{-0.1cm}
    g_i = \sum_{j=0}^{B -1} h_{i\times B + j} \times 2^j
\end{equation}
where $g_i$ is the desired conductance of the $i^{th}$ device representing a weight. Note that negative weights are mapped in a similar manner.
Considering the impact of device variations, the actually programmed conductance value $gp_i$ is as follows:
\vspace{-0.2cm}
\begin{equation}
    \vspace{-0.1cm}
    gp_i = g_i + \Delta{g}
\end{equation}
where $\Delta{g}$ is the deviation from the desired conductance value $g_i$. 

Thus when weight is programmed, the actual value $\mathcal{W}_{p}$ mapped on the devices would be: 
\vspace{-0.2cm}
\begin{align}
\vspace{-0.2cm}
\begin{split}
    \mathcal{W}_{p}     & =\frac{\max{|\mathcal{W}|}}{2^H - 1} \sum_{i=0}^{H/B -1}2^{i\times B}{gp_i } \\
                        & = \mathcal{W}_{des} + \frac{\max{|\mathcal{W}|}}{2^H - 1} \sum_{i=0}^{H/B-1}{\Delta{g} \times 2^{i\times B}}\\
\end{split}
\vspace{0.0 cm}
\end{align}

To simulate the above process, we follow the settings consistent with existing works. Specifically, we set $B=2$ based on existing works~\cite{jiang2020device, yan2022swim}, while $H$ is specified by each model. For the device variation model, we adopt $\Delta{g} \sim \mathcal{N}(0, \sigma_d)$ (if not specified), which indicates that $\Delta{g}$ follows Gaussian distribution with a mean of zero and a standard deviation of $\sigma_d$. We constrain $\sigma_d \leq 0.4$ as this is a reasonable range that can be realized by device-level optimizations such as write-verify based on the measurement results. Our model and parameter settings are in line with that of RRAM devices reported in~\cite{shim2020two}. 




\subsection{Experimental Setup}
\textbf{Platforms and Metrics}: All experiments are conducted on PyTorch using an on-the-shelf GPU. To precisely capture the performance (accuracy) of the DNN model under device variations, our report data points are averaged from 5 identical runs. For the evaluation metric, if not specified, we report the $1^{st}$ percentile accuracy, which is \KPP~using accuracy as the performance metric and with $k=1$. To obtain the \KPP~of a DNN model under sufficiently high precision, we choose to run 10,000 Monte Carlo simulations ($N_{sample}$ = 10,000). Since our experiments show that 10,000 runs can output $1^{st}$ percentile accuracy whose 95\% confidence interval is $\pm 0.009$ based on the central limit theorem. 

\textbf{Baselines for Comparison}: We compare TRICE with three baselines that are built upon training: (1) training w/o noise injection, (2) CorrectNet~\cite{eldebiky2022correctnet}, and (3) injecting Gaussian noise in training~\cite{jiang2020device, yang2022tolerating}. For a fair comparison, we do not compare TRICE with other orthogonal methods like NAS-based DNN topologies design~\cite{yan2021uncertainty, jiang2020device} or Bayesian Neural Networks~\cite{gao2021bayesian}, given TRICE can be used together with them. 

\textbf{Hyperparameters Setting}: For all experiments, TRICE uses the same hyperparameter setups: $start = 0$, $end = 2\times\sigma_d$, $th = 2$, $ep=100$, $warm=5$ and $N_{train}=300$, where $\sigma_d$ is the standard deviation for device variation. We limit the range of $\sigma_d$ as suggested by Sect.~\ref{sect:model} and report the effectiveness of TRICES across different $\sigma_d$ values within that range. For other training hyperparameters such as learning rate, batch size, and learning rate schedulers, we follow the best practice in training a noise-free model. 

\subsection{The Effectiveness of TRICE on MNIST Dataset}


\begin{table}[ht]
    \centering
    \caption{Effectiveness of TRICE method on model LeNet for MNIST across different $\sigma_d$ values. The performance is shown in $1^{st}$ percentile accuracy. The baselines are vanilla DNN training w/o noise injection, CorrectNet~\cite{eldebiky2022correctnet}, and injecting Gaussian noise in training~\cite{jiang2020device,yang2022tolerating
    }. Injecting RC-Gaussian noise with hand-picked hyperparameters (RC-Manual) is also shown as an ablation study.}
    \vspace{-0.1cm}
    \begin{tabular}{cccccc}
        \toprule
        Dev. var.  &\multicolumn{5}{c}{Training Method}\\
        ($\sigma_d$)    & w/o noise & CorrectNet    & Gauss.  & RC-Manual & TRICE \\
        \midrule
        0.00    & 99.01     & 97.99         & 98.86     & 98.88     & 98.94 \\
        0.05    & 93.31     & 97.56         & 97.45     & 96.89     & \textbf{98.08} \\
        0.10    & 70.72     & 90.66         & 95.59     & 95.47     & \textbf{95.99} \\
        0.15    & 38.15     & 67.70         & 87.60     & 90.43     & \textbf{90.58} \\
        0.20    & 19.81     & 39.54         & 66.04     & 75.47     & \textbf{77.82} \\
        0.25    & 11.95     & 22.26         & 40.27     & 50.14     & \textbf{54.12} \\
        0.30    & 08.58     & 14.26         & 23.09     & 28.56     & \textbf{38.51} \\
        0.35    & 06.89     & 10.83         & 14.38     & 16.83     & \textbf{25.29} \\
        0.40    & 06.05     & 09.23         & 10.38     & 11.61     & \textbf{17.94} \\
        \bottomrule
    \end{tabular}
    \label{tab:GvPT}
    \vspace{-0.4cm}
\end{table}

We first compare TRICE with the aforementioned baselines using the model LeNet to recognize the 10-class handwritten digits dataset MNIST~\cite{lecun1998gradient}. LeNet is a plain convolutional neural network consisting of two convolution layers and three fully connected layers. All weights and layer outputs (\emph{i.e.}, activations) are quantized to four bits ($H=4$). We also compare TRICE with injecting right-censored Gaussian noise with handpicked hyperparameters (RC-Manual) as an ablation study. Table~\ref{tab:GvPT} shows the $1^{st}$ percentile accuracy of models trained with different training methods under different levels of device variations ($\sigma_d$) following the noise model discussed in Section~\ref{sect:model}. As shown in Table~\ref{tab:GvPT}, compared with training w/o noise, CorrectNet improves the $1^{st}$ percentile accuracy by up to 19.94\%, but this is not comparable to the improvement of up to 49.44\% by injecting Gaussian noise and up to 58.01\% by our proposed TRICE. We can also observe that, compared with injecting Gaussian noise, TRICE can improve the $1^{st}$ percentile accuracy by up to 15.42\%. It is clear that TRICE outperforms all baselines in generating models with higher $1^{st}$ percentile accuracy in all simulated $\sigma_d$ values. Moreover, TRICE demonstrates more significant improvement when facing large device variations while still delivering comparable accuracy when $\sigma_d$ is too small to distinguish the difference between different training methods. Because CorrectNet cannot generate a model with higher robustness compared with injecting Gaussian noise, we do not show the results for it in the latter experiments. The ablation study also shows that TRICE outperforms injection right-censored Gaussian with handpicked hyperparameters (RC-Manual) and the improvement in $1^{st}$ percentile accuracy is up to 9.95\%, so we do not show the result of RC-Manual in the remainder of this paper.

\begin{figure*}
    \vspace{-0.4cm}
    \centering
    \begin{minipage}[b]{0.44\linewidth}
        \includegraphics[trim=10 25 40 50, clip, width=1.\linewidth]{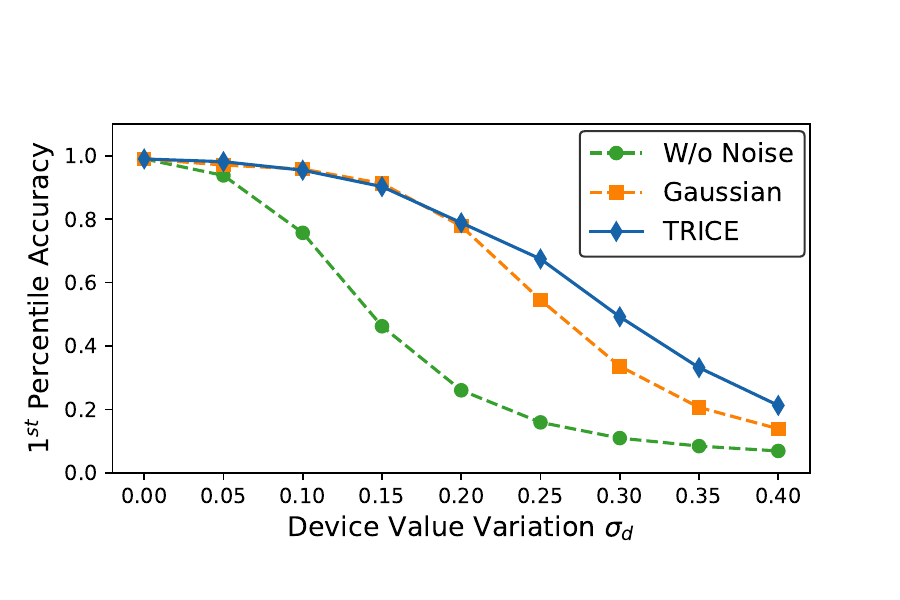}
        \vspace{-0.5cm}
        \subcaption{FeFET$_1$}
        \label{fig:H4}
    \end{minipage}
    \hspace{0.03\linewidth}
    \begin{minipage}[b]{0.44\linewidth}
        \includegraphics[trim=10 25 40 50, clip, width=1.\linewidth]{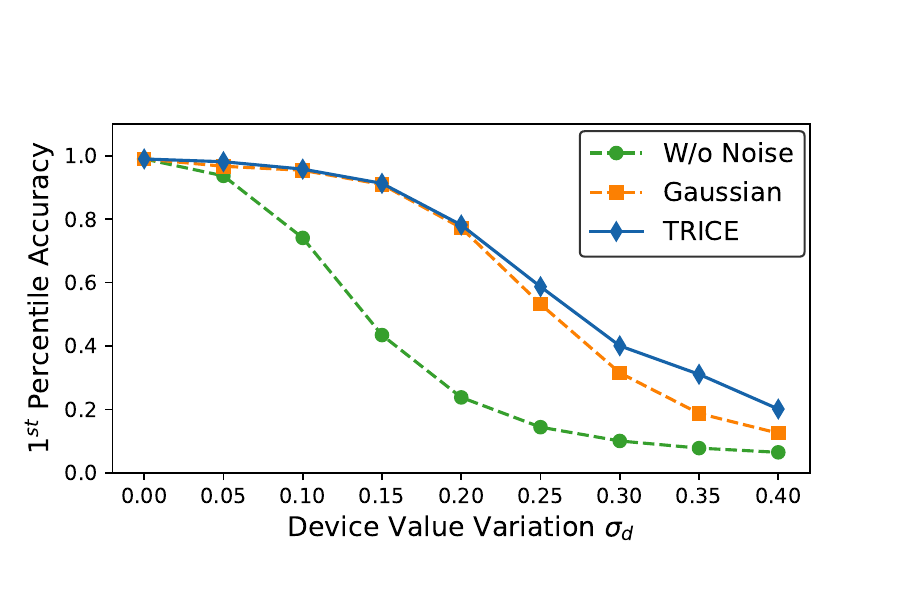}  
        \vspace{-0.5cm}
        \subcaption{FeFET$_2$}
        \label{fig:H2}
    \end{minipage}
    \vspace{-0.2cm}
    \caption{Comparison of the $1^{st}$ percentile accuracy achieved by models trained using TRICE and baseline methods on LeNet for dataset MNIST targeting devices (a) FeFET$_1$ and (b) FeFET$_2$. The x-axis represents the magnitude of device value variation ($\sigma_d$) and the y-axis represents the $1^{st}$ percentile accuracy.}
    \vspace{-0.7cm}
\end{figure*}

\subsection{The Effectiveness of TRICE in Large Models}

After showing the effectiveness of TRICE in a small model LeNet for MNIST, here we further demonstrate the effectiveness of TRICE by comparing it with the baselines in larger DNN models for larger datasets. We choose two representative models VGG-8~\cite{simonyan2014very} and ResNet-18~\cite{he2016deep}. Both models use a 6-bit quantization ($H=6$) for weights and activations. They both perform image classification tasks for dataset CIFAR-10~\cite{krizhevsky2009learning}. 
As shown in Fig.~\ref{fig:cifar} and Fig.~\ref{fig:res18}, 
TRICE clearly outperforms all baselines in most device value deviation values and performs similarly as baselines in some rare cases where device value deviation is too small to make an impact or too large to perform a valid classification. Compared with injecting Gaussian noise, 
TRICE improves the $1^{st}$ percentile accuracy by up to 25.09\%, and 26.01\% in VGG-8 for CIFAR-10 and ResNet-18 for CIFAR-10, respectively.



\subsection{The Effectiveness of TRICE in Different Devices}

To demonstrate the scalability of TRICE, we also show the effectiveness of TRICE on NVCiM platforms using different types of NVM devices. As discussed in Section~\ref{sect:model}, previous experiments use a four level (2-bit, $B=2$) device as in~\cite{jiang2020device, yan2022swim}. More specifically, it is a four-level RRAM device whose device value deviation model is $\Delta{g} \sim \mathcal{N}(0, \sigma_d)$, which means $\Delta{g}$ follows Gaussian distribution with a mean of zero and a standard deviation of $\sigma_d$, independent of the programmed device conductance.

We further analyze the effectiveness of TRICE on two real-world FeFET devices whose device value deviation magnitude varies as its programmed conductance changes. Their device models are derived from measurement results in~\cite{chakraborty2020beol}. Specifically, a generalized device value variation model for a four-level device is:
\vspace{-0.1cm}
\begin{equation}
    \vspace{-0.1cm}
    \begin{aligned}
        \begin{array}{ll}
            gp_i &= g_i + \Delta{g} \\
            \Delta{g} &\sim \mathcal{N}(0, \sigma_h)
        \end{array}, \ \ \ \  
        \sigma_h = & \begin{cases}
            \sigma_{d0},\ \ \  if\  g_i = 0\\
            \sigma_{d1},\ \ \  if\  g_i = 1\\
            \sigma_{d2},\ \ \  if\  g_i = 2\\
            \sigma_{d3},\ \ \  if\  g_i = 3\\
        \end{cases}
    \end{aligned}
\end{equation}
which means $\Delta{g}$ follows Gaussian distribution with a mean of zero and a standard deviation of $\sigma_h$ but the $\sigma_h$ value differs as its programmed conductance changes. We abstract the behaviors of the two FeFET devices to be:
\vspace{-0.1cm}
\begin{align}
\vspace{-0.1cm}
    \text{FeFET}_1 \rightarrow \{\sigma_{d0} = \sigma_{d3} = \sigma_d, \sigma_{d1} = \sigma_{d2} = 4\sigma_d\}\label{eq:fe1} \\
    \text{FeFET}_2 \rightarrow \{\sigma_{d0} = \sigma_{d3} = \sigma_d, \sigma_{d1} = \sigma_{d2} = 2\sigma_d\}\label{eq:fe2}
\end{align}

\begin{figure}[b]
    \vspace{-0.6cm}
    \centering
    \includegraphics[trim=0 0 0 0, clip, width=1\linewidth]{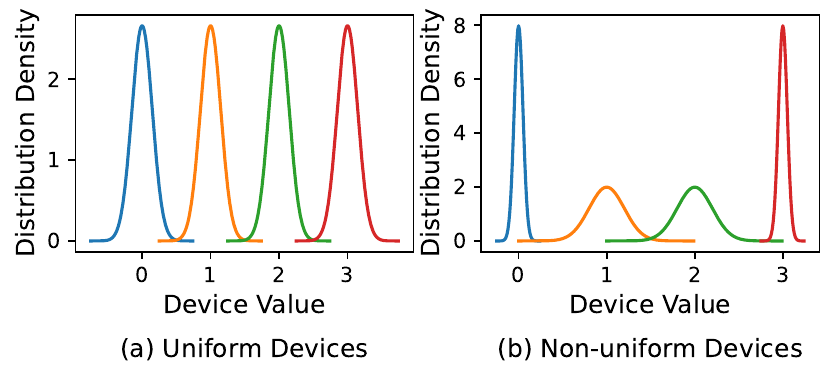}
    \vspace{-0.7cm}
    \caption{Illustration of uniform and non-uniform devices. (a) Uniform devices suffer from the same magnitude of noise when programmed to different conductance values. (b) Non-uniform devices suffer from different magnitudes of noise when programmed to different conductance values. The perturbation is more significant when the conductance value is 1 and 2.}
    \vspace{-0.2cm}
    \label{fig:devices}
\end{figure}
This means the devices suffer from more device variations when they are programmed to value 1 and 2 and suffer from less device variations when they are programmed to value 0 and 3. As a comparsion, we show the conductance ($gp$) distribution of the previously used RRAM device and FeFET$_2$ in Fig.~\ref{fig:devices}a and  Fig.~\ref{fig:devices}b, respectively.

We report the effectiveness of TRICE in NVCiM platforms using FeFET$_1$ and FeFET$_2$ in Fig.~\ref{fig:H4} and Fig.~\ref{fig:H2}, respectively. As expected, again, it is obvious that TRICE outperforms all baselines in most $\sigma_d$ values and performs similarly as baselines where device value deviation is too small to make an impact. Compared with injecting Gaussian noise, TRICE improves the $1^{st}$ percentile accuracy by up to 15.61\%, and 12.34\% in FeFET$_1$ and FeFET$_2$, respectively.

\subsection{Ablation Study for Different Noise Candidates}


\begin{figure}[ht]
    \vspace{-0.5cm}
    \centering
    \includegraphics[trim=0 0 0 0, clip, width=0.9\linewidth]{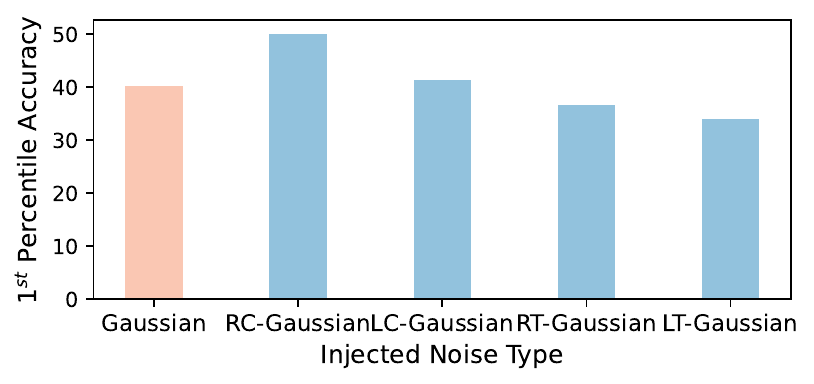}
    \vspace{-0.2cm}
    \caption{Comparison of injecting different types of noise in training LeNet for MNIST. The y-axis represents the $1^{st}$ percentile accuracy of models trained by injecting different types of noise when $\sigma_d = 0.25$.}
    \label{fig:RCVAll}
    \vspace{-0.2cm}
\end{figure}

We also show the effectiveness of injecting RC-Gaussian noise in the training process by comparing it against injecting other three noise candidates: LC-Gaussian, RT-Gaussian, and LT-Gaussian noise. Here the result of training with Gaussian noise is also included as a baseline. Without loss of generality, we perform this study on the LeNet for MNIST dataset using uniform RRAM devices with $\sigma_d = 0.25$. As shown in Fig.~\ref{fig:RCVAll}, training with RC-Gaussian noise shows a clear advantage over training with other types of noise by at least 8.76\%. Note that training with left and right truncated Gaussian performs even worse than injecting Gaussian noise because they exhibit lower accuracy w/o the presence of device variations.
\section{Conclusions}\label{sect:conclusion}
In this work, we propose to use k-th percentile performance (\KPP) instead of widely used average performance as a metric to evaluate the realistic worst-case performance of a DNN model. By analyzing the properties of DNN models and noise injection-based training, we show that the conventional Gaussian noise injection training is far from optimal in improving \KPP. Thus, we propose TRICE which injects right-censored noise during training. Extensive experiments show that TRICE clearly outperforms SOTA baselines in improving the k-th percentile performance of DNN models.

\clearpage
\bibliographystyle{ieeetr}
\bibliography{M7_References}

\begin{thebibliography}{10}

\bibitem{ramesh2021zero}
A.~Ramesh, M.~Pavlov, G.~Goh, S.~Gray, C.~Voss, A.~Radford, M.~Chen, and
  I.~Sutskever, ``Zero-shot text-to-image generation,'' in {\em International
  Conference on Machine Learning}, pp.~8821--8831, PMLR, 2021.

\bibitem{brown2020language}
T.~Brown, B.~Mann, N.~Ryder, M.~Subbiah, J.~D. Kaplan, P.~Dhariwal,
  A.~Neelakantan, P.~Shyam, G.~Sastry, A.~Askell, {\em et~al.}, ``Language
  models are few-shot learners,'' {\em Advances in neural information
  processing systems}, vol.~33, pp.~1877--1901, 2020.

\bibitem{jiang2020device}
W.~Jiang, Q.~Lou, Z.~Yan, L.~Yang, J.~Hu, X.~S. Hu, and Y.~Shi,
  ``Device-circuit-architecture co-exploration for computing-in-memory neural
  accelerators,'' {\em IEEE Transactions on Computers}, vol.~70, no.~4,
  pp.~595--605, 2020.

\bibitem{chen2016eyeriss}
Y.-H. Chen, J.~Emer, and V.~Sze, ``Eyeriss: A spatial architecture for
  energy-efficient dataflow for convolutional neural networks,'' {\em ACM
  SIGARCH computer architecture news}, vol.~44, no.~3, pp.~367--379, 2016.

\bibitem{yan2020single}
Z.~Yan, Y.~Shi, W.~Liao, M.~Hashimoto, X.~Zhou, and C.~Zhuo, ``When single
  event upset meets deep neural networks: Observations, explorations, and
  remedies,'' in {\em 2020 25th Asia and South Pacific Design Automation
  Conference (ASP-DAC)}, pp.~163--168, IEEE, 2020.

\bibitem{shafiee2016isaac}
A.~Shafiee, A.~Nag, N.~Muralimanohar, R.~Balasubramonian, J.~P. Strachan,
  M.~Hu, R.~S. Williams, and V.~Srikumar, ``Isaac: A convolutional neural
  network accelerator with in-situ analog arithmetic in crossbars,'' {\em ACM
  SIGARCH Computer Architecture News}, vol.~44, no.~3, pp.~14--26, 2016.

\bibitem{shim2020two}
W.~Shim, J.-s. Seo, and S.~Yu, ``Two-step write--verify scheme and impact of
  the read noise in multilevel rram-based inference engine,'' {\em
  Semiconductor Science and Technology}, vol.~35, no.~11, p.~115026, 2020.

\bibitem{yan2021uncertainty}
Z.~Yan, D.-C. Juan, X.~S. Hu, and Y.~Shi, ``Uncertainty modeling of emerging
  device based computing-in-memory neural accelerators with application to
  neural architecture search,'' in {\em 2021 26th Asia and South Pacific Design
  Automation Conference (ASP-DAC)}, pp.~859--864, IEEE, 2021.

\bibitem{yan2022radars}
Z.~Yan, W.~Jiang, X.~S. Hu, and Y.~Shi, ``Radars: Memory efficient
  reinforcement learning aided differentiable neural architecture search,'' in
  {\em 2022 27th Asia and South Pacific Design Automation Conference
  (ASP-DAC)}, pp.~128--133, IEEE, 2022.

\bibitem{gao2021bayesian}
D.~Gao, Q.~Huang, G.~L. Zhang, X.~Yin, B.~Li, U.~Schlichtmann, and C.~Zhuo,
  ``Bayesian inference based robust computing on memristor crossbar,'' in {\em
  2021 58th ACM/IEEE Design Automation Conference (DAC)}, pp.~121--126, IEEE,
  2021.

\bibitem{he2019noise}
Z.~He, J.~Lin, R.~Ewetz, J.-S. Yuan, and D.~Fan, ``Noise injection adaption:
  End-to-end reram crossbar non-ideal effect adaption for neural network
  mapping,'' in {\em Proceedings of the 56th Annual Design Automation
  Conference 2019}, pp.~1--6, 2019.

\bibitem{yang2022tolerating}
X.~Yang, C.~Wu, M.~Li, and Y.~Chen, ``Tolerating noise effects in
  processing-in-memory systems for neural networks: A hardware--software
  codesign perspective,'' {\em Advanced Intelligent Systems}, vol.~4, no.~8,
  p.~2200029, 2022.

\bibitem{yan2022computing}
Z.~Yan, X.~S. Hu, and Y.~Shi, ``Computing-in-memory neural network accelerators
  for safety-critical systems: Can small device variations be disastrous?,'' in
  {\em Proceedings of the 41st IEEE/ACM International Conference on
  Computer-Aided Design}, pp.~1--9, 2022.

\bibitem{feinberg2018making}
B.~Feinberg, S.~Wang, and E.~Ipek, ``Making memristive neural network
  accelerators reliable,'' in {\em 2018 IEEE International Symposium on High
  Performance Computer Architecture (HPCA)}, pp.~52--65, IEEE, 2018.

\bibitem{liu2019fault}
T.~Liu, W.~Wen, L.~Jiang, Y.~Wang, C.~Yang, and G.~Quan, ``A fault-tolerant
  neural network architecture,'' in {\em 2019 56th ACM/IEEE Design Automation
  Conference (DAC)}, pp.~1--6, IEEE, 2019.

\bibitem{wu2020adversarial}
D.~Wu, S.-T. Xia, and Y.~Wang, ``Adversarial weight perturbation helps robust
  generalization,'' {\em Advances in Neural Information Processing Systems},
  vol.~33, pp.~2958--2969, 2020.

\bibitem{tsai2021formalizing}
Y.-L. Tsai, C.-Y. Hsu, C.-M. Yu, and P.-Y. Chen, ``Formalizing generalization
  and adversarial robustness of neural networks to weight perturbations,'' {\em
  Advances in Neural Information Processing Systems}, vol.~34, 2021.

\bibitem{peng2019dnn+}
X.~Peng, S.~Huang, Y.~Luo, X.~Sun, and S.~Yu, ``Dnn+ neurosim: An end-to-end
  benchmarking framework for compute-in-memory accelerators with versatile
  device technologies,'' in {\em 2019 IEEE international electron devices
  meeting (IEDM)}, pp.~32--5, IEEE, 2019.

\bibitem{eldebiky2022correctnet}
A.~Eldebiky, G.~L. Zhang, G.~Boecherer, B.~Li, and U.~Schlichtmann,
  ``Correctnet: Robustness enhancement of analog in-memory computing for neural
  networks by error suppression and compensation,'' {\em Design, Automation and
  Test in Europe Conference (DATE) 2023}, 2023.

\bibitem{chen2021pruning}
C.-Y. Chen and K.~Chakrabarty, ``Pruning of deep neural networks for
  fault-tolerant memristor-based accelerators,'' in {\em 2021 58th ACM/IEEE
  Design Automation Conference (DAC)}, pp.~889--894, IEEE, 2021.

\bibitem{yao2020fully}
P.~Yao, H.~Wu, B.~Gao, J.~Tang, Q.~Zhang, W.~Zhang, J.~J. Yang, and H.~Qian,
  ``Fully hardware-implemented memristor convolutional neural network,'' {\em
  Nature}, vol.~577, no.~7792, pp.~641--646, 2020.

\bibitem{shin2021fault}
H.~Shin, M.~Kang, and L.-S. Kim, ``Fault-free: A fault-resilient deep neural
  network accelerator based on realistic reram devices,'' in {\em 2021 58th
  ACM/IEEE Design Automation Conference (DAC)}, pp.~1039--1044, IEEE, 2021.

\bibitem{jeong2022variation}
S.~Jeong, J.~Kim, M.~Jeong, and Y.~Lee, ``Variation-tolerant and low r-ratio
  compute-in-memory reram macro with capacitive ternary mac operation,'' {\em
  IEEE Transactions on Circuits and Systems I: Regular Papers}, 2022.

\bibitem{yan2022swim}
Z.~Yan, X.~S. Hu, and Y.~Shi, ``Swim: Selective write-verify for
  computing-in-memory neural accelerators,'' in {\em 2022 59th ACM/IEEE Design
  Automation Conference (DAC)}, IEEE, 2022.

\bibitem{dangel2020backpack}
F.~Dangel, F.~Kunstner, and P.~Hennig, ``Backpack: Packing more into
  backprop,'' in {\em International Conference on Learning Representations},
  2020.

\bibitem{lecun1998gradient}
Y.~LeCun, L.~Bottou, Y.~Bengio, and P.~Haffner, ``Gradient-based learning
  applied to document recognition,'' {\em Proceedings of the IEEE}, vol.~86,
  no.~11, pp.~2278--2324, 1998.

\bibitem{simonyan2014very}
K.~Simonyan and A.~Zisserman, ``Very deep convolutional networks for
  large-scale image recognition,'' {\em arXiv preprint arXiv:1409.1556}, 2014.

\bibitem{he2016deep}
K.~He, X.~Zhang, S.~Ren, and J.~Sun, ``Deep residual learning for image
  recognition,'' in {\em Proceedings of the IEEE conference on computer vision
  and pattern recognition}, pp.~770--778, 2016.

\bibitem{krizhevsky2009learning}
A.~Krizhevsky, G.~Hinton, {\em et~al.}, ``Learning multiple layers of features
  from tiny images,'' 2009.

\bibitem{chakraborty2020beol}
W.~Chakraborty, B.~Grisafe, H.~Ye, I.~Lightcap, K.~Ni, and S.~Datta, ``Beol
  compatible dual-gate ultra thin-body w-doped indium-oxide transistor with
  ion= 370$\mu$a/$\mu$m, ss= 73mv/dec and ion/ioff ratio> 4$\times$ 109,'' in
  {\em 2020 IEEE Symposium on VLSI Technology}, pp.~1--2, IEEE, 2020.

\end{thebibliography}

\end{document}